# From Image Latent Space to Fuzzy Rules: Interpretable Analysis of Gastrointestinal Foundation Model

Michael D. Vasilakakis[1][0000-0003-0739-0040] and Dimitris K. Iakovidis[1][0000-0002-5027-5323]

[1] Department of Computer Science and Biomedical Informatics, University of Thessaly, Lamia, Greece
vasilaka@uth.gr, diakovidis@uth.gr

**Abstract.** Foundation models pretrained on large-scale datasets demonstrate strong transferability to medical imaging tasks. However, understanding how their latent representations encode clinically relevant information remains an open challenge in safety-critical domains. This study proposes a prototype-based fuzzy-rule framework that interprets the patch-level features produced by the inner layers of pretrained foundation models, without any fine-tuning. Class-specific prototypes are learned by clustering in the feature space, yielding compact visual patterns. Patch features are then expressed as prototype similarities and classified by fuzzy rules with linguistic IF-THEN conditions that are human readable. The framework is applied across the final two blocks of ViT-S/16 backbones pretrained on ImageNet-1K and GastroNet-5M, and benchmarked against k-nearest neighbours, kernel SVM, and linear probing under identical frozen features, on wireless capsule endoscopy classification, gastrointestinal endoscopy classification, and colonic polyp segmentation. The experimental analysis shows that the proposed method, without backbone fine-tuning, reaches accuracy comparable to these black-box classifiers, and that domain-specific pretraining yields features that are both discriminative and symbolically compressible. Because the resulting rules are extracted from real data and expressed in interpretable terms, they are further used as an instrument to investigate synthetic medical images, providing a human-readable account of which real prototypes and rules a generator reproduces or fails to reproduce, localising where a synthetic image departs from real tissue rather than summarising it with a single score. The framework thus offers a transparent, depth-resolved view of how foundation models organise clinically relevant structure, together with a practical downstream use of the extracted rules.

## 1 Introduction

Foundation models have significantly advanced medical image analysis through large-scale pretraining and improved transferability across downstream tasks. In gastrointestinal (GI) endoscopy, models are commonly pretrained on natural image datasets such as ImageNet-1K [1], whose visual statistics differ substantially from endoscopic imagery. Recent studies demonstrate that domain-specific pretraining on large medical datasets, such as GastroNet-5M, substantially improves downstream performance, especially when combined with self-supervised learning strategies[1–3]. Despite these advances, understanding how pretrained models internally organize clinically relevant structures within their features, and how that organisation changes across network depth, remains an open challenge in safety-critical domains where transparent decision-making is essential.

Deep architectures such as Vision Transformers (ViT) [4, 5] produce high-dimensional latent embeddings whose discriminative organization and class separability are difficult to interpret. Post hoc explanation techniques, including saliency maps and Grad-CAM[6, 7], provide localized visualizations but often fail to characterize the structural geometry of the learned representation space. Moreover, saliency-based explanations can be misleading, reflecting model biases rather than causal relationships between inputs and outputs [8]. Inherently interpretable approaches analyze model behavior through structured reasoning. Prototype-based models explain predictions through similarity to representative examples, revealing class-dependent structure in latent space [8, 9]. Complementarily, fuzzy rule-based systems express classification outcomes using linguistic if-then conditions that align naturally with human reasoning[10–12], offering a transparent alternative to black-box decision mechanisms.

This study proposes a prototype-based fuzzy-rule framework that operates on the patch-level features extracted from the inner layers of pretrained foundation models, without any fine-tuning. Class-specific prototypes are learned by clustering in the feature space, giving compact visual patterns. Patch features are then transformed into prototype-similarity features and classified by sparse fuzzy rules with linguistic IF-THEN conditions designed for human inspection. The aim is not to design an interpretable classifier for accuracy alone, but to characterise what interpretable structure is already present in the features produced by a given pretrained backbone, and how that structure varies across inner layers and pretraining domain. To this end the framework is swept across the final two blocks of ViT-S/16 under both ImageNet-1K and GastroNet-5M pretraining, and its rule-based readout is compared against standard classifiers, namely k-nearest neighbours, kernel SVM, and linear probing, evaluated on the same features extracted using the pre-trained models under identical conditions.

The experimental analysis serves first to establish that this interpretability has no significant cost in performance. On wireless capsule endoscopy classification (KID) [13], GI endoscopy classification (Kvasir-v2)[14], and colonic polyp segmentation (Kvasir-SEG) [15], the proposed method, without backbone fine-tuning, reaches accuracy comparable to the black-box classifiers while, unlike them, yielding an explicit and compact set of human-readable rules. Classification is examined both for each abnormality against normal tissue and in a pooled normal-versus-abnormal setting that reflects a first-line screening decision. Across depths and pretraining sources, domain-specific pretraining tends to produce latent spaces that are both discriminative and symbolically compressible, so that fewer and shorter rules describe the same structure.

Because the extracted rules are derived from real data and expressed in interpretable terms, they can be repurposed beyond classification. This study uses them as an instrument to investigate synthetic medical images: fitted only on real data, the rule base reads a synthetic image through the vocabulary of real tissue and reports, in human-readable terms, which real prototypes and rules a generator over-activates or fails to activate (fidelity). It names the specific prototypes and rules where a generator diverges from real tissue, an explanation that aggregate generative metrics do not provide. This investigation shows that fidelity and utility need not coincide, and it names the prototypes and rules responsible for the gap.

The contributions of this study are as follows. First, it presents an interpretable prototype-based fuzzy-rule framework that reads the inner layers of frozen pretrained foundation models and turns their patch-level features into compact, human-inspectable IF-THEN rules. Second, through a depth-resolved and pretraining-resolved analysis it shows that this readout, obtained without backbone fine-tuning, is comparable in accuracy to black-box classifiers while remaining interpretable, and that domain-specific pretraining yields representations that are simultaneously discriminative and symbolically compressible. Third, it illustrates a downstream use of the extracted rules as an interpretable fidelity auditor of synthetic medical images, naming in clinically meaningful terms the prototypes and rules a generator fails to reproduce.

## 2 Related Work

**Prototype-based interpretable classification**. Case-based prototype networks make predictions by comparing an image to learned prototypical parts. ProtoPNet [16] introduced the "this looks like that" paradigm; TesNet[17] structured the prototype space on a Grassmann manifold; PIP-Net[18] added self-supervised prototype alignment and a non-negative, additive head; ProtoTree [19] organized prototypes in a decision tree. These methods are typically trained end-to-end, coupling prototype quality to task-specific backbone adaptation. These methods jointly optimize the backbone alongside the prototype layer, so prototype quality and backbone adaptation cannot be disentangled. This study instead operates on non-fine-tuned features, so the analysis reflects the representation itself rather than a co-adapted head, and it examines how the interpretable structure changes across inner layers and pretraining domain rather than at a single depth. Because a single centroid is a poor summary of a visually diverse class, several prototypes per class are retained rather than one, which matters in particular when abnormalities are pooled into a single, heterogeneous abnormal class. For the same reason a numerical comparison against end-to-end prototype networks is not made, since it would conflate feature quality with the effect of backbone adaptation; the benchmark is instead against classifiers that share the same frozen features.

**Post-hoc prototype interpretability**. Closer to our regime, recent work extracts prototype-based explanations from already-trained models without backbone retraining; In [20] the authors decompose a trained classification head into part-prototypes post-hoc while preserving the original accuracy, and interpretable-by-design approaches over trained foundation features have been proposed for general representations [8, 9]. In particular, the framework [8] formalizes interpretable classification over frozen foundation-model feature spaces, the setting adopted here. This study shares the same gradient-free premise but differs in explanation modality as characterizing similarity through linguistic fuzzy sets rather than visual heatmaps, and in scope, by systematically comparing feature quality across trained layers rather than at a single depth.

**Fuzzy rule-based interpretability**. Fuzzy rule systems express decisions as linguistic IF–THEN conditions aligned with human reasoning, and their interpretability is well studied, including cognitive bounds on rule length[21]. Such systems have primarily characterized structured/tabular attributes[10, 12]; recent work quantifies prototype similarity for vision[18]. In particular, the rule-construction mechanism used here adapts the framework[12], which was developed for structured/tabular attributes, to frozen ViT patch features. Building on this, the fuzzy rules used here operate on the graded magnitude of prototype similarity through linguistic membership functions, retaining a notion of partial match, and are derived directly from features across trained layers rather than from a trained architecture. This graded, partial-match view also makes the rules suitable for the downstream analysis of synthetic data, where a tolerant description is preferable to a hard decision.

**Evaluating synthetic medical images**. Since part of this study repurposes the extracted rules to examine synthetic images, it also relates to the evaluation of generative models. Synthetic-data quality is commonly summarised by distributional distances and embedding-based precision and recall, by the separability of a real-versus-synthetic discriminator, and by downstream train-on-synthetic, test-on-real protocols, interpretable set-level utility measures have recently been introduced for the medical domain[22]. These provide a single score and treat the feature space as a black box. The rule base used here is complementary, giving a per-prototype and per-rule account of where synthetic and real data diverge while remaining human-readable.

# 3 Methodology

## 3.1 Latent Feature Extraction and Prototype Construction

Let $I = \{i_1\ , i_2, \dots, i_N\}$ be a set of $N$ of training input images. Each training image carries a single image-level label; under a weak-supervision assumption this label is broadcast to all of the image's patches, giving patch-level labels $\mathrm{y}_n = (y^1, \dots, y^G)$, where $y^g \in \{0, .., C\}$ indicates the class, normal or abnormal tissue, of patch $g$. For the segmentation data the patch labels are instead derived from the pixel-level masks (strong supervision), which avoids the weak-label assumption that, for the classification tasks, also marks normal and background patches within an abnormal image as abnormal and may bias the abnormal prototypes toward image-level context rather than lesion-specific tissue. For ViT-s/16, each image $i_n$ is partitioned into $G$ non-overlapping patches (G=196, on a 14×14 grid). Each patch is embedded through the pretrained transformer encoder (ViT-S/16 pretrained with DINO self-supervision and used without any fine-tuning), producing contextualized latent feature vectors $\boldsymbol{F_n} = (f_1, \dots, f_G)$ where $f_g \in \mathbb{R}^d, g = 1, \dots, G, d = 384$ represents the latent embedding of patch $g$.
For each class $c$, latent patch features belonging to that class are grouped using K-medoids clustering. The resulting cluster centers define a vocabulary of class-specific prototypes $P = \{p_1, \dots, p_K\}, k = 1, \dots, K$. Each prototype $p_k$ is associated with a class label $y_k$. These prototypes represent typical latent patterns of normal and abnormal tis sue in the pretrained feature space. Each prototype corresponds to a representative latent embedding derived from real image patches and therefore preserves semantic correspondence with observable tissue patterns (Fig.1).

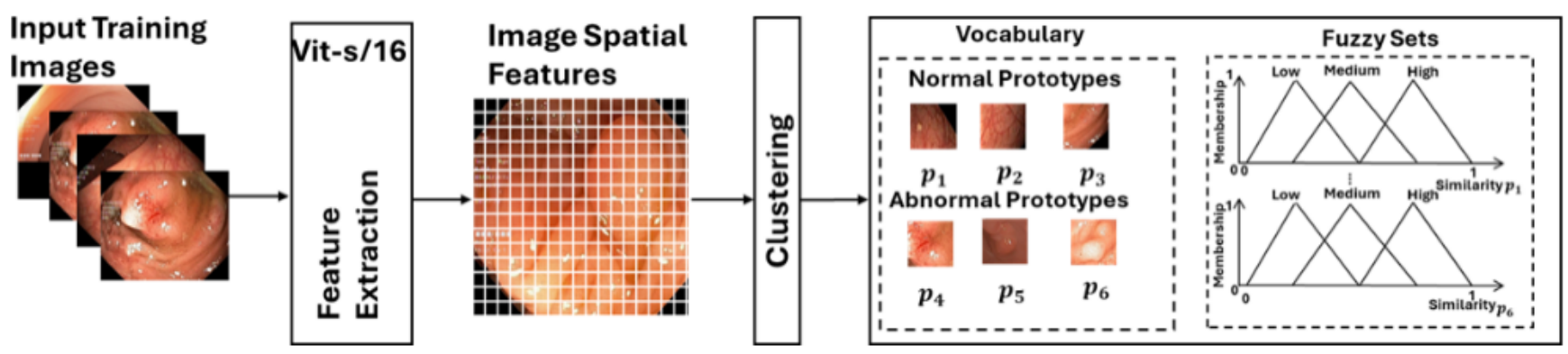

**Fig. 1.** Prototype vocabulary construction and fuzzy membership functions for normal (p1–p3) and abnormal (p4–p6) tissue classes extracted from ViT-s/16 spatial features.

### 3.2 Prototype-Based Latent Representation

Each feature vector $f_g$ is encoded based on its similarity to prototypes in $P$. The similarity $s_{g,p_k}$ is computed as

$$s_{g,p_k} = 1 - \frac{\|f_g - p_k\|}{\sum_{p_j \in P} \|f_g - p_j\|} \quad (1)$$

This vocabulary-wide normalization reduces cross-class interference and reflects the idea that human perception classifies new instances by comparing them against familiar prototypes. The resulting values are then rescaled by a per-prototype min–max normalisation fitted on the training patches, so that each similarity $s_{g,p_k} \in [0,1]$, with validation and test values clipped to this range. Each patch is therefore represented by a similarity vector as $s_g = (s_{g,p_1}, \dots, s_{g,p_K})$.To enhance the interpretability of the computed similarities, linguistic descriptions are used, which can be easily comprehended by humans. For each prototype $p_k$, similarity values of training patches are clustered into $L$ groups using one-dimensional k-means. The cluster center of each group is denoted as $a_{p_k,l} \in [0,1], l = 1, .., L$ , and the centers are sorted in ascending order. For each cluster center $a_{p_k,l}$, a fuzzy set $\tilde{A}_{p_k,l}$ is defined. Each fuzzy set is associated with a membership function $\mu_{p_k,l}(s_{g,p_k}) \in [0,1]$, typically a triangular membership function. Thus, numerical similarities are translated into linguistic similarity features (e.g., Low,…, High), providing a structured and interpretable description of the latent space, as illustrated in Fig. 1, where normal and abnormal prototypes and their associated fuzzy membership functions are shown.

### 3.3 Interpretable Classification

Let $\boldsymbol{F}$ denote the set of latent feature vectors extracted from an unknown test image. For each patch feature $f_g$, similarities $s_{g,p_k}$ are computed. Each similarity is assigned to its dominant linguistic label $l' = \underset{l\in[1,L]}{\operatorname{argmax}}\left(\mu_{p_k,l}(s_{g,p_k})\right)$. Let $P_c = \{p_k \epsilon P | y_k = c\}$ denote the set of prototypes belonging to class $c$. A class-level fuzzy evidence for feature vector $f_g$ is defined as $E_g^c = \frac{1}{|P_c|}\sum_{p_k \in P_c} s_{g,p_k} \cdot \mu_{p_k,l'}(s_{g,p_k})$, where the normalization by $|P_c|$ ensures comparability across classes with different prototype counts. The product $s_{g,p_k}\mu_{p_k,l'}(s_{g,p_k})$ jointly encodes both the geometric proximity of a patch to a prototype in the latent space and the linguistic confidence of that proximity, ensuring that prototypes which are both close to the patch and strongly activated by the dominant fuzzy label contribute most to the class evidence, while distant or linguistically ambiguous prototypes are naturally downweighted. Given two distinct classes $c \neq c'$ a binary fuzzy rule is defined as

$$Rule^{c,c'}: If \;\; E_g^c > E_g^{c'}, \text{then classify } f_g \text{ in class } c \tag{2}$$

A relative certainty measure is defined as

$$y^{c,c'}(f_g) = E_g^c - E_g^{c'} \tag{3}$$

where $y^{c,c'}(f_g) \in [-1,1]$. A positive value indicates stronger fuzzy evidence for class $c$, while a negative value favor class $c'$. This formulation allows numerical decisions to remain consistent with their linguistic rule representation, ensuring agreement between symbolic explanations and classification outcomes (Fig.2). In case of multiple classes maximum average relative certainty of all pairs of classes is computed $y^c(f_g) = \frac{1}{C}\sum_{c'=1,c\neq c'}^{C} y^{c,c'}(f_g)$. The class with the highest average relative certainty is assigned as the final prediction. At the image level, the per-patch predictions are aggregated by majority voting over the G patches, an image being labelled abnormal when the number of patches predicted abnormal exceeds a fixed threshold; for segmentation, the same per-patch decisions are assembled into a lesion mask.

Although the class-level fuzzy evidence $E_g^c$ aggregates contributions from all prototypes $P_c'$, explicit fuzzy rules are constructed at the linguistic level to preserve interpretability. For each feature vector $f_g$ and class $c$, the dominant fuzzy label for every prototype $p_k \in P_c'$ is first determined as $l' = \underset{l\in[1,L]}{\operatorname{argmax}}\left(\mu_{p_k,l}(s_{g,p_k})\right)$. However, retaining every prototype's condition can produce an overly long IF-part. According to [21], a rule remains interpretable when the number of conditions in its IF-part does not exceed 7±2, in line with the limits of human comprehension. Therefore, for each class only the top $M^c$ conditions are kept, so that the IF-part length is bounded and depends on the number of linguistic terms rather than on the number of prototypes per class. For example, in Fig. 2 the rule set is reduced to a single dominant condition per class ($M^c = 1$), and the patch-level predictions are aggregated into a segmentation mask.

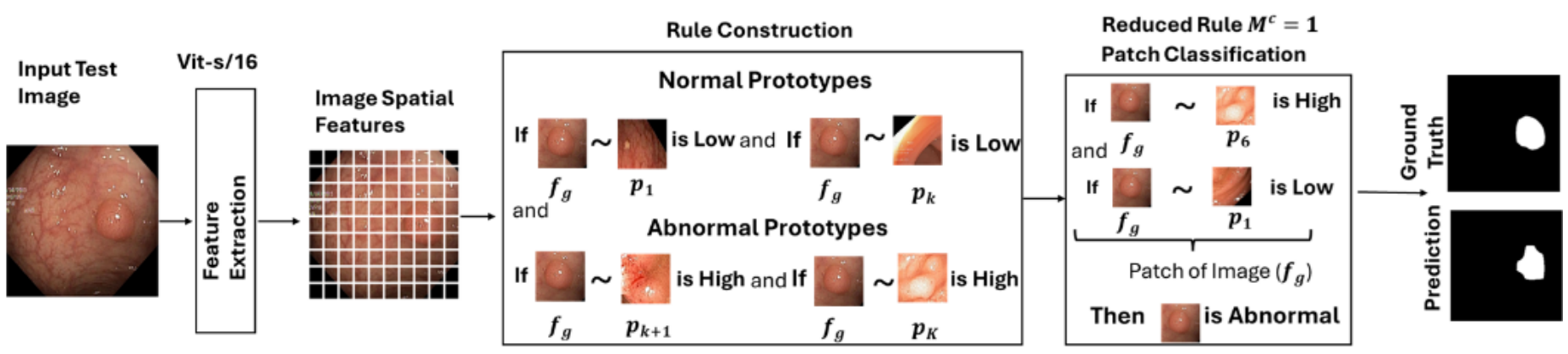

**Fig. 2.** Fuzzy rule construction and reduction ($M^c = 1$) for patch-level classification and lesion mask prediction, with corresponding ground truth.

The linguistic rules are assembled once, from the training patches only, and are thereafter held fixed. For every training patch feature $f_g$ and every prototype $p_k \in P_c$, the dominant term $l' = argmax_{l \in [1,L]}(\mu_{p_k,l}(s_{g,p_k}))$ is recorded together with the sign of its contribution to the class evidence, where a term supporting class c is marked (+) and one opposing it (−). Each patch thus induces a signed pattern over the (class, prototype, term) slots, and after the $M^c$ reduction this pattern has at most $M^c$ conditions per class. Collecting these patterns over all training patches and removing duplicates yields the rule base $R = \{r_1, \ldots, r_R\}$, a compact set of IF-THEN rules of the form of Eq. (2) whose conditions are expressed as Class: Term with a (+)/(−) sign. A representative extracted pair, with its prototype crops and relative-certainty values, is shown and analysed in Section 4.3 (Fig. 3). Because the rules are read off the same fixed evidence $E_g^c$ used for classification, they reproduce the classifier's decisions rather than defining a new one, and each prototype $p_k$ in a condition can be displayed as its nearest real training patch, so a condition such as Abnormal: *High points to concrete tissue*. This rule base, obtained from the training split without any optimization, is later reused unchanged to describe unseen real images and to audit synthetic ones.

# 4 Experiments and results

## 4.1 Experimental setup

Three gastrointestinal benchmarks are used. Wireless capsule endoscopy classification uses Dataset 2 of the KID database, which consists of normal images together with inflammatory, vascular, and polypoid lesions, reflecting clinical class imbalance. The gastrointestinal endoscopy classification uses Kvasir-v2. The colonic polyp segmentation uses Kvasir-SEG, the only task with pixel-level ground truth and therefore real patch labels. Patch features are read from a ViT-S/16 encoder trained with DINO self-supervision and used without any fine-tuning, under two pretraining sources, GastroNet-5M (GN-5M) and ImageNet-1K (IN-1K), and at two depths, the final block (block 11) and the preceding block (block 10). Each image yields 196 patch tokens of dimension 384 on a 14×14 grid. Data are split at the image level with a stratified 40 percent test partition and results are averaged over ten random repeats. For the classification tasks the image-level label is broadcast to all patches during vocabulary fitting, whereas Kvasir-SEG uses mask-derived patch labels.

Per-patch decisions are produced by the fuzzy rule of Eq. (2) and aggregated into an image label by majority vote over the $G = 196$ patches, an image being labelled abnormal when at least half of its patches are predicted abnormal. This count-based aggregation of patch predictions into an image-level decision is the standard weakly-supervised, multiple-instance setting[23]. The proposed fuzzy framework (PF) is compared against three baselines that share the same pretrained ViT-S/16 features under this identical aggregation harness: a kernel SVM, a linear probe and a prototype k-nearest-neighbour classifier (kNN). Proposed Framework (PF) is evaluated over prototypes, with the number of prototypes per class swept over $\{5, 7, 9, 11, 13, 15\}$. The fuzzy vocabulary uses per-class k-means prototypes with triangular membership functions and euclidean similarity; its number of linguistic terms L is swept over {3, 5, 7, 9} and selected by grid search on a held-out validation split, with L=3 (terms Low, Med, High) and the reduced phrase chosen for the image-level classification tasks. Classification is reported with AUC[24] and segmentation with per-patch Dice[3] at the 14x14 patch resolution.

### 4.2 Classification and Segmentation Performance

The primary classification setting groups all abnormal categories into a single class, matching a first-line screening decision in which the question is whether an image is normal or contains any abnormality. **Tables 1** and **2** give this overall result on KID and Kvasir-v2 at both depths. On the domain backbone GastroNet-5m (GN-5M) the fuzzy framework is close to the two trained probes, within a few AUC points, and at or above the k-NN. On Kvasir-v2 it stays near 0.99 and clears k-NN comfortably. The gap to the probes widens only under ImageNet (IN-1K) pretraining, where the prototype-based methods lose more from the heterogeneous pooled class, but the fuzzy framework remains ahead of k-NN there as well. The two depths give essentially the same picture.

Table 3 breaks the abnormal side into its individual categories against normal tissue at both depths. Of the three abnormal categories in KID, the per-category analysis focuses on the inflammatory and polypoid lesions, as representative chronic-inflammatory and neoplastic findings respectively; the vascular category is subsumed within the overall abnormal-versus-normal setting of Table 1 and Table 2.

**Table 1.** KID, overall abnormal versus normal classification (AUC), for ViT-S/16 Block 10 (B10) and Block 11 (B11)

| **Method** | **GN-5M·B10** | **IN-1K·B10** | **GN-5M ·B11** | **IN-1K·B11** |
|---|---|---|---|---|
| PF | 0.80±0.01 | 0.78±0.01 | 0.80±0.01 | 0.77±0.01 |
| SVM-RBF | 0.83±0.01 | 0.84±0.01 | 0.83±0.01 | 0.84±0.01 |
| Linear probe | 0.83±0.01 | 0.84±0.01 | 0.83±0.01 | 0.83±0.01 |
| kNN | 0.80±0.01 | 0.77±0.01 | 0.79±0.01 | 0.76±0.01 |

**Table 2.** Kvasir-v2, overall abnormal versus normal classification (AUC), for ViT-S/16 Block 10 (B10) and Block 11 (B11)

| **Method** | **GN-5M·B10** | **IN-1K·B10** | **GN-5M ·B11** | **IN-1K·B11** |
|---|---|---|---|---|
| PF | 0.99±0.01 | 0.94±0.01 | 0.99±0.01 | 0.93±0.01 |
| SVM-RBF | 0.99±0.01 | 0.99±0.01 | 0.99±0.01 | 0.99±0.01 |
| Linear probe | 0.99±0.01 | 0.97±0.01 | 0.99±0.01 | 0.96±0.01 |
| kNN | 0.96±0.01 | 0.87±0.01 | 0.98±0.01 | 0.84±0.01 |

The ranking matches the overall setting: the two trained probes lead, and the fuzzy framework follows within roughly two to seven AUC points depending on category, narrowest on the inflammatory lesions and widest on polypoid under ImageNet features. It stays ahead of the k-nearest-neighbour control in every column, and the two depths give essentially the same picture.

On Kvasir-SEG, which has real patch-level labels, the fuzzy framework uses the reduced phrase, the same configuration as in the classification tasks and its better-performing setting here. The two trained probes lead on Dice, but only narrowly: on the domain backbone the fuzzy framework reaches 0.819 Dice against about 0.84 for the probes (Table 4), a gap of roughly two points, while remaining clearly ahead of the k-nearest-neighbour control and being the only method that yields an explicit rule set of fixed length $M^C$. The domain backbone (GN-5M) dominates ImageNet (IN-1K) for every method, and the penultimate block matches or slightly exceeds the final block.

Because the reduced phrase makes the rule length depend only on the number of linguistic terms and not on the number of prototypes, the proposed framework can be made more accurate by adding prototypes without lengthening its rules. Table 5 sweeps both the prototype count $P$ and the number of linguistic terms $L$ on the same task, alongside the k-NN control at matched prototype count. At fixed L, PF AUC rises with the number of prototypes, from 0.78 at five prototypes to 0.85 at fifteen, while the rule length stays constant (it is $M^c$, independent of $P$), and PF exceeds the k-NN control at every matched count. This shows the advantage over the nearest-prototype control is consistent rather than a single-point effect, and it is the mechanism behind the comparability reported above: accuracy can be raised through $P$ while the explanation length is untouched. The number of linguistic terms L trades accuracy against interpretability in the other direction: L=3 is best at almost every prototype count for the image-level classification tasks (the sole exception being $P$=7), and since the rule length is $M^c$, larger L also lengthens every rule, so L=3 is retained as the setting that is both most accurate and most compact. The segmentation task, which has real patch labels, benefits from larger L.

**Table 3.** Per-abnormality classification (AUC)

| Method | Block | KID infl· GN-5M | KID polyp· GN-5M | KID infl·IN | KID polyp· IN-1K |
|---|---|---|---|---|---|
| PF | B10 | 0.84±0.01 | 0.86±0.02 | 0.82±0.01 | 0.83±0.03 |
| PF | B11 | 0.83±0.01 | 0.86±0.02 | 0.82±0.02 | 0.81±0.03 |
| SVM | B10 | 0.88±0.01 | 0.91±0.02 | 0.89±0.01 | 0.89±0.04 |
| SVM | B11 | 0.88±0.01 | 0.92±0.02 | 0.88±0.02 | 0.88±0.04 |
| Linear probe | B10 | 0.87±0.01 | 0.92±0.02 | 0.88±0.01 | 0.86±0.05 |
| Linear probe | B11 | 0.87±0.01 | 0.93±0.02 | 0.86±0.01 | 0.85±0.04 |
| kNN | B10 | 0.82±0.01 | 0.83±0.03 | 0.79±0.02 | 0.80±0.04 |
| kNN | B11 | 0.81±0.02 | 0.82±0.02 | 0.76±0.04 | 0.79±0.05 |

**Table 4.** Kvasir-SEG segmentation Dice

| Method | GN-5M·B10 | IN-1K·B10 | GN-5M ·B11 | IN-1K·B11 |
|---|---|---|---|---|
| PF | 0.82±0.01 | 0.64±0.01 | 0.81±0.01 | 0.63±0.01 |
| SVM-RBF | 0.84±0.01 | 0.72±0.01 | 0.84±0.01 | 0.71±0.01 |
| Linear probe | 0.84±0.01 | 0.71±0.01 | 0.84±0.01 | 0.71±0.01 |
| kNN | 0.80±0.01 | 0.57±0.01 | 0.79±0.01 | 0.56±0.01 |

**Table 5.** Fuzzy AUC over prototypes per class $P$ and linguistic terms $L$ (KID, GN-5M, block 11), with the k-NN reference at matched prototype count. The reduced-phrase rule length is $M^c$ and is independent of $P$.

| $P$ | $L$=3 | L=5 | $L$=7 | $L$=9 | kNN (best k) |
|---|---|---|---|---|---|
| 5 | 0.778 | 0.766 | 0.787 | 0.683 | 0.645 |
| 7 | 0.792 | 0.794 | 0.763 | 0.778 | 0.757 |
| 9 | 0.833 | 0.797 | 0.782 | 0.789 | 0.816 |
| 11 | 0.834 | 0.804 | 0.787 | 0.774 | 0.823 |
| 13 | 0.843 | 0.841 | 0.814 | 0.814 | 0.811 |
| 15 | 0.847 | 0.831 | 0.821 | 0.804 | 0.823 |

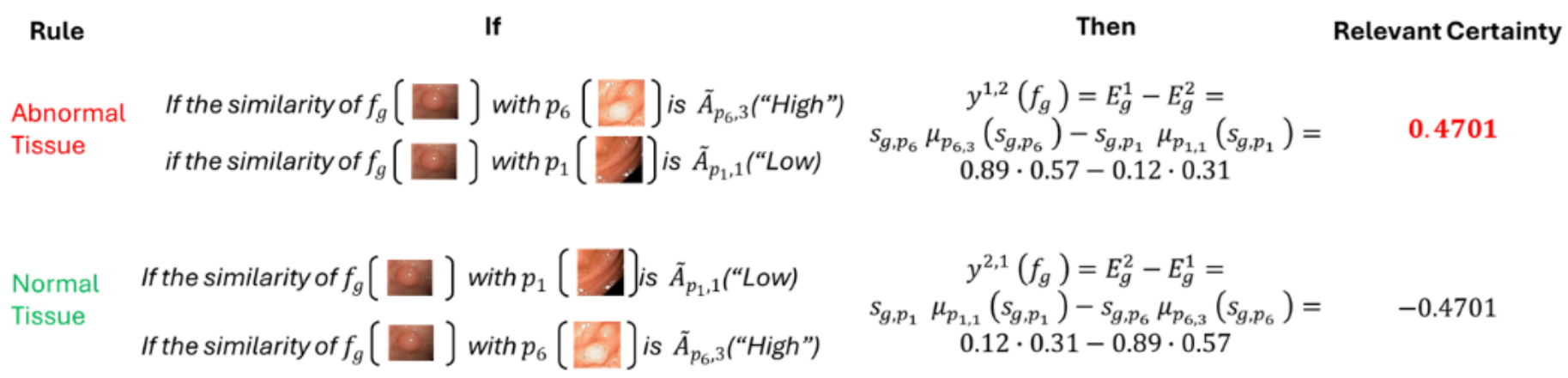

**Fig. 3.** Representative rule pair extracted by the proposed framework for an abnormal patch $f_g$ (Kvasir-SEG, GastroNet-5M ViT-S/16, block 10), over the abnormal prototype p6 and the normal prototype p1. Each IF-part condition is a Class:Term atom; the Abnormal rule (p6 High, p1 Low) gives relative certainty $y = E^{Abnormal} - E^{Normal} = 0.4701$ and the symmetric Normal rule gives −0.4701. Prototypes are shown as their nearest real training-patch crop.

### 4.3 Interpretability Analysis

Only the fuzzy framework yields an explicit rule representation. From the real KID training split (GN-5M, block 10) the framework extracts a compact rule base of 35 rules (17 predicting Normal, 18 predicting Abnormal), several with only two conditions in their IF-part and none exceeding six, so the whole base sits within the 7±2 comprehension bound. Because each prototype is displayed as its nearest real training patch, a condition such as Abnormal:High points to concrete tissue rather than an abstract index. As a soundness check, scoring the extracted rules reproduces the reported classifier decisions exactly, so the rules explain the reported numbers rather than defining a separate model.

Fig. 3 shows a representative rule pair from the Kvasir-SEG vocabulary, anticipated in Section 3.3, for a single patch $f_g$. The IF-part of each rule references prototype crops with their linguistic similarity terms, so the decision can be read both visually and in words. For the Abnormal rule, $f_g$ is highly similar to the abnormal prototype p6 (High) and weakly similar to the normal prototype p1 (Low), yielding a positive relative certainty of 0.4701 that assigns the patch to the Abnormal class; the symmetric Normal rule produces the counterpart −0.4701, so the two rules agree. The numerical THEN-part exposes the decision margin, so that the prototype vocabulary and the linguistic terms can in principle be inspected and adjusted without retraining the backbone. The same kind of patch-level rule-reading illustrated in Fig. 3 is what the following section turns on synthetic images: reading a generated patch through its own real vocabulary and comparing how often each rule fires is precisely how the fidelity of a generator is assessed.

### 4.4 Synthetic-image audit via fuzzy rule fidelity

The rule base extracted from the real training split, obtained without any optimization, is reused unchanged as an instrument to read synthetic images. Three generators are audited, StyleGAN[25], TiDE[26], and TiDE II[27]. For each synthetic image the same pipeline is applied as for real images, the same reading illustrated in Fig. 3, mapping its patches to prototype similarities, then to dominant linguistic terms, and then to the IF-THEN rules, and the per-patch rule-firing and prototype-activation frequencies are averaged over a set of sampled synthetic images and compared against those of a real reference set of the same class. Each table reports two such frequencies: rule-firing frequency, the fraction of an image's 196 patches on which a given abnormal rule fires, and prototype-activation frequency, the fraction of patches whose nearest lesion-class prototype is a given prototype. A faithful generator matches the real column on both; where it deviates, the audit names which prototypes it over- or under-activates and which rules it fails to fire, so the discrepancy is described in concrete, human-readable terms instead of one number.

**Table 6.** Rule-level fidelity of synthetic polyps against real polyps (KID, GN-5M, block 10). Values are means over each image set (real N=44; StyleGAN and TiDE N=150 sampled synthetic polyps each); rows are per-patch fractions except the first. Prototypes are shown in Fig. 4.

| Signal | real polyp | StyleGAN | TiDE |
|---|---|---|---|
| Mean abnormal-patch vote (of 196) of image set | 124.00 | 85.00 | 99.90 |
| Polyp rule R1 firing | 0.70 | 0.61 | 0.77 |
| Polyp rule R3 firing | 0.10 | 0.01 | 0.03 |
| Polyp Prototype p1 activation | 0.23 | 0.37 | 0.24 |
| Polyp Prototype p4 activation | 0.24 | 0.22 | 0.28 |
| Polyp Prototype p6 activation | 0.10 | 0.00 | 0.01 |

**Table 7.** Rule-level fidelity of synthetic ulcers against real ulcers (Kvasir-v2, GN-5M, block 10), the faithful contrast to the polyp case. Values are means over each image set (real N=200 held-out ulcer images; each generator N=150 sampled synthetic ulcers); rows are per-patch fractions except the first. Prototypes are shown in Fig. 4.

| Signal | real ulcer | StyleGAN | TiDE | TiDE II |
|---|---|---|---|---|
| Average abnormal-patch vote (of 196) of image set | 132.30 | 141.20 | 117.60 | 148.40 |
| Ulcer rule R3 firing | 0.35 | 0.40 | 0.47 | 0.53 |
| Ulcer rule R1 firing | 0.30 | 0.38 | 0.22 | 0.30 |
| Ulcer Prototype p3 activation | 0.27 | 0.27 | 0.24 | 0.20 |
| Ulcer Prototype p4 activation | 0.31 | 0.29 | 0.38 | 0.38 |

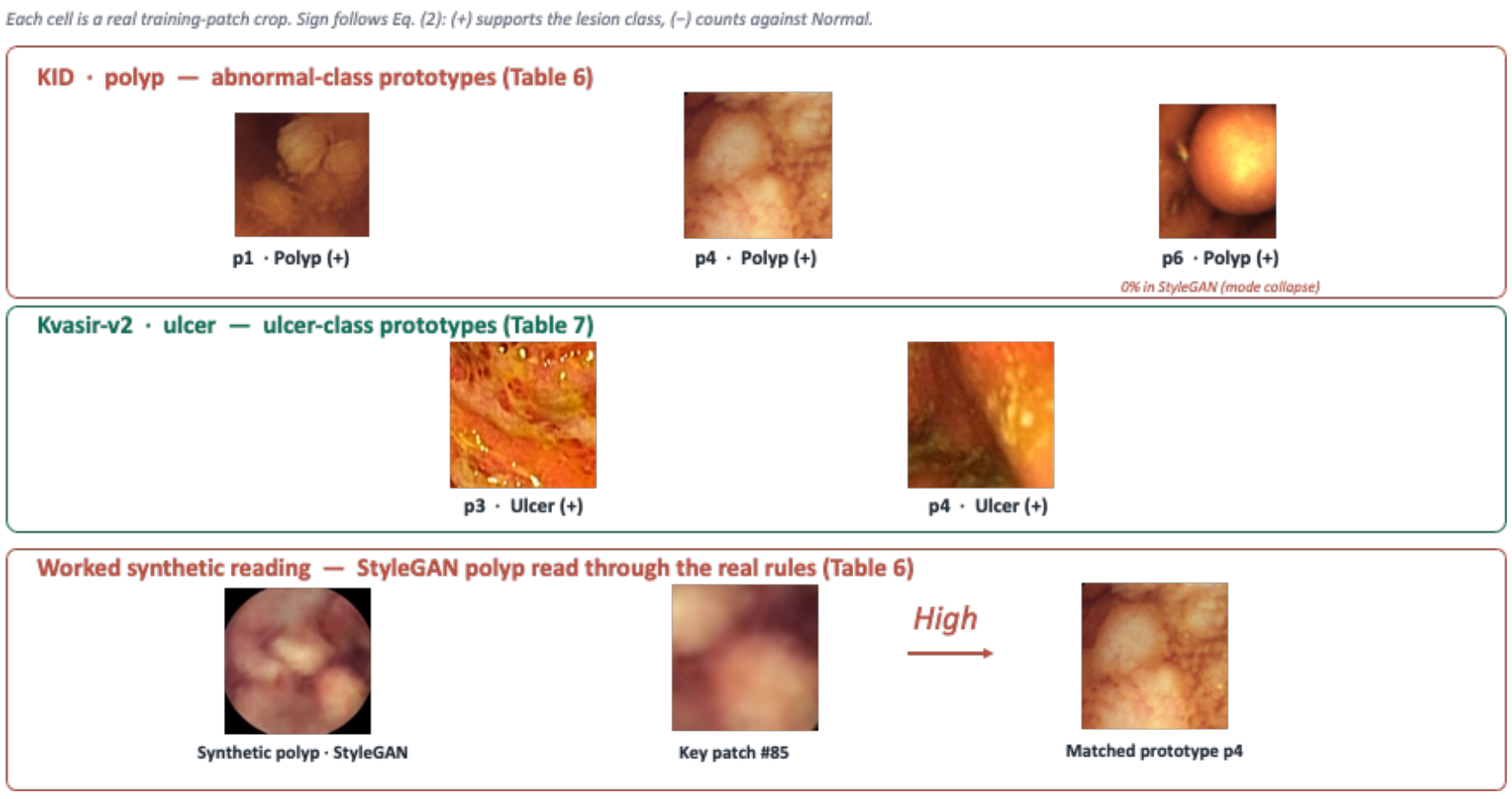

**Fig. 4** Audit prototypes and a worked synthetic reading (GastroNet-5M, block 10). Each prototype is shown as its nearest real training-patch crop, and its sign follows Eq. (2): a match to a lesion-class prototype enters a rule as (+) evidence for that class and (−) against Normal. Bottom: one StyleGAN polyp read through the real rules, showing the synthetic image, the key patch (#85), and the real prototype it matches at High similarity (p4).

The audit separates faithful from unfaithful generators in concrete terms. On KID, the synthetic polyps are read as less abnormal than real ones: averaged over all 44 real polyp images and 150 sampled synthetic polyps per generator, the mean abnormal-patch vote drops from 124 of 196 for real polyps to 85 for StyleGAN, the confident abnormal rule R1 fires less often, and abnormal prototype p6 is never activated at all, falling from 10% of real polyp patches to 0%, a localised and nameable mode collapse (Table 6). This is the specific, inspectable reason the real rules find the StyleGAN polyps less convincing, rather than a scalar distance that would leave the cause unstated.

The ulcer case is the faithful counterpart. On Kvasir-v2 the synthetic ulcers reproduce the real ulcer signature closely: averaged over 200 held-out real ulcer images (disjoint from those used to fit the vocabulary) and 150 sampled synthetic ulcers per generator, the abnormal-patch vote is comparable to real (132 for real against 141 for StyleGAN), and the dominant rules and prototypes match, so the rule base cannot readily separate the synthetic ulcers from real ones (Table 7). The contrast between the two cases shows that the instrument does not simply flag all synthetic data as different. It distinguishes a generator that reproduces the clinical rule signature from one that does not, and states which signal fails in each case.

Fig. 4 makes these readings concrete. Its top and middle panels show the prototypes named in Tables 6 and 7 as their nearest real training-patch crops, so a condition such as Abnormal:p6 or Ulcer:p3 points to identifiable tissue rather than an index, and the sign attached to each follows Eq. (2). Its bottom panel walks through one StyleGAN polyp: the synthetic image, the single most abnormal patch, and the real prototype that patch matches at High similarity (p4). The same reading, applied over all 196 patches and averaged across the image set, is what produces the frequencies in Table 6, so the figure and the table describe the same operation at the level of one patch and of the whole set respectively

## 5 Conclusion

This study presented an interpretable prototype-rule framework that reads the inner layers of a pretrained foundation model and turns their patch-level features into compact, human-inspectable IF-THEN rules, without any fine-tuning of the backbone. Prototypes are learned by clustering real patch features, their similarities are described by linguistic terms, and the resulting rules classify each patch while remaining short enough to fall within the 7±2 comprehension budget, with each condition grounded in a concrete real-tissue crop.

Across wireless capsule endoscopy classification, gastrointestinal endoscopy classification, and polyp segmentation, evaluated on ViT-S/16 features under both ImageNet-1K and GastroNet-5M pretraining and at two depths, the experimental analysis showed that this interpretability is achieved with no significant cost in the classification accuracy. The rule-based readout, obtained without backbone fine-tuning, is comparable to the trained SVM and linear probes, trailing them by a few AUC points on classification and by about two Dice points on the domain backbone for segmentation, and it is ahead of the k-nearest-neighbour. Two consistent patterns emerged from the depth- and pretraining-resolved study: domain-specific pretraining produces latent spaces that are both discriminative and amenable to prototype-based reasoning, and the penultimate block matches or slightly exceeds the final block, indicating that clinically relevant structure is not confined to the deepest representation. Because the reduced phrase fixes the rule length independently of the number of prototypes, accuracy can be improved by adding prototypes without lengthening the explanation.

Beyond classification, the study illustrated that the same rule base, fit only on real data, can be turned on synthetic images to give a rule-level, human-readable account of how a generator's images depart from real tissue, naming the specific prototypes and rules that a synthetic image fails to reproduce. The worked examples showed the instrument pointing to a concrete, localised failure in one case and to close agreement in another, which suggests that interpretable rules can serve not only to explain a classifier but also to inspect generated data in clinically meaningful terms.

Future work will extend the framework along several directions. A natural next step is to evaluate the extracted rules with clinicians, assessing how readily the prototype and rule explanations are understood and used in practice. The synthetic-image audit can be broadened from the illustrative single-image cases shown here to an aggregate study across many generated images and additional generators, turning the rule-level reading into a systematic fidelity instrument.

**Acknowledgments.** This work is part of the European project SEARCH, which is supported by the Innovative Health Initiative Joint Undertaking (IHI JU) under grant agreement No. 101172997. The JU receives support from the European Union's Horizon Europe research and innovation programme and COCIR, EFPIA, Europa Bio, MedTech Europe, Vaccines Europe, Medical Values GmbH, Corsano Health BV, Syntheticus AG, Maggioli SpA, Motilent Ltd, Ubitech Ltd, Hemex Benelux, Hellenic Healthcare Group, German Oncology Center, Byte Solutions Unlimited, AdaptIT GmbH. Views and opinions expressed are however those of the author(s) only and do not necessarily reflect those of the aforementioned parties. Neither of the aforementioned parties can be held responsible for them.

**Disclosure of Interests.** The authors have no competing interests to declare that are relevant to the content of this article.